\documentclass[letterpaper, 10 pt, conference]{ieeeconf}

\IEEEoverridecommandlockouts                              

\usepackage{graphicx}
\usepackage{caption}
\usepackage{amsmath,amssymb,enumerate}

\usepackage{amsthm}
\usepackage{mathtools}
\usepackage{breqn}
\usepackage{algorithm, algorithmicx, algpseudocode}

\usepackage{blindtext}
\usepackage{gensymb}
\usepackage{xparse}
\usepackage{lipsum}
\usepackage{mathrsfs}
\usepackage[mathscr]{euscript}
\usepackage{times}
\usepackage{cite} 
\usepackage{multicol}
\usepackage[caption=false,font=footnotesize]{subfig}
\usepackage{amsfonts}
\usepackage[utf8]{inputenc}
\usepackage[T1]{fontenc}
\usepackage{textcomp}
\usepackage{amsfonts}
\usepackage{soul}
\usepackage{multirow}
\usepackage{booktabs}
\usepackage[usenames,dvipsnames,svgnames,table, xcdraw]{xcolor}

\usepackage[colorlinks=true,pdfpagemode=UseNone,citecolor=black,linkcolor=black,urlcolor=BrickRed]{hyperref}

\newtheorem{problem}{Problem}

\DeclareMathOperator*{\argmax}{arg\,max}

\renewcommand{\thefigure}{\arabic{figure}}

\newcommand{\squeezeup}{\vspace{-3mm}}

\newcommand{\Hcal}{\mathcal{H}}
\newcommand{\Ical}{\mathcal{I}}

\DeclareDocumentCommand{\vectorToSkew}{ O{} }{\left(#1\right)_\times}

\title{\LARGE \bf
Adaptive Continuous Visual Odometry from RGB-D Images
}

\author{Tzu-Yuan Lin, William Clark, Ryan M. Eustice, Jessy W. Grizzle \\ Anthony Bloch, and Maani Ghaffari
\thanks{*This work was partially supported by the Toyota Research Institute (TRI), partly under award number N021515. Funding for J. Grizzle was in part provided by TRI and in part by NSF Award No.~1808051. Funding for W. Clark and A. Bloch was provided by NSF DMS 1613819 and AFSOR FA9550-18-10028.}
\thanks{T.Y.~Lin, W.~Clark, R.~Eustice, J. Grizzle, A. Bloch, and M.~Ghaffari are with the University of Michigan, Ann Arbor, MI 48109, USA. {\tt\small\{tzuyuan, wiclark, eustice, grizzle, abloch, maanigj\}@umich.edu}.}%
}

\begin{document}

\maketitle
\thispagestyle{empty}
\pagestyle{empty}

\begin{abstract}
In this paper, we extend the recently developed continuous visual odometry framework for RGB-D cameras to an adaptive framework via online hyperparameter learning. We focus on the case of isotropic kernels with a scalar as the length-scale. In practice and as expected, the length-scale has remarkable impacts on the performance of the original framework. Previously it was handled using a fixed set of conditions within the solver to reduce the length-scale as the algorithm reaches a local minimum. We automate this process by a greedy gradient descent step at each iteration to find the next-best length-scale. Furthermore, to handle failure cases in the gradient descent step where the gradient is not well-behaved, such as the absence of structure or texture in the scene, we use a search interval for the length-scale and guide it gradually toward the smaller values. This latter strategy reverts the adaptive framework to the original setup. The experimental evaluations using publicly available RGB-D benchmarks show the proposed adaptive continuous visual odometry outperforms the original framework and the current state-of-the-art. We also make the software for the developed algorithm publicly available.

\end{abstract}

\section{INTRODUCTION}

Visual odometry using depth cameras is the problem of finding a rigid-body transformation between two colored point clouds. This problem arises frequently in robotics and computer vision and is an integral part of many autonomous systems~\cite{whelan2013robust,scherer2013efficient,valenti2014autonomous,loianno2015cooperative,huang2017visual}. Direct visual odometry methods minimize the photometric error using image intensity values measured by the camera~\cite{audras2011real,newcombe2011dtam,steinbrucker2011real}. The explicit representation between color information and 2D/3D geometry (image or Euclidean space coordinates) is not directly available; hence, the current direct methods use numerical differentiation for computing the gradient and are limited to fixed image size and resolution, given the camera model and measurements for re-projection of the 3D points. In this setup, a coarse-to-fine image pyramid~\cite[Section 3.5]{szeliski2010computer} is constructed to solve the same problem several times with initialization provided by solving the previous coarser step.

Alternatively, the Continuous Visual Odometry (CVO) is a continuous and direct formulation and solution for the \mbox{RGB-D} visual odometry problem~\cite{MGhaffari-RSS-19}. Due to the continuous representation of CVO, it neither requires the association between two measurement sets nor the same number of measurements within each set. In addition, there is no need for constructing a coarse-to-fine image pyramid in the continuous sensor registration framework developed in~\cite{MGhaffari-RSS-19}. In this framework, the joint appearance and geometric embedding is modeled by representing the processes (\mbox{RGB-D} images) in a Reproducing Kernel Hilbert Space (RKHS)~\cite{scholkopf2001generalized,berlinet2004reproducing}. 

\begin{figure}[t]
    \centering
    \subfloat{\includegraphics[width=\columnwidth]{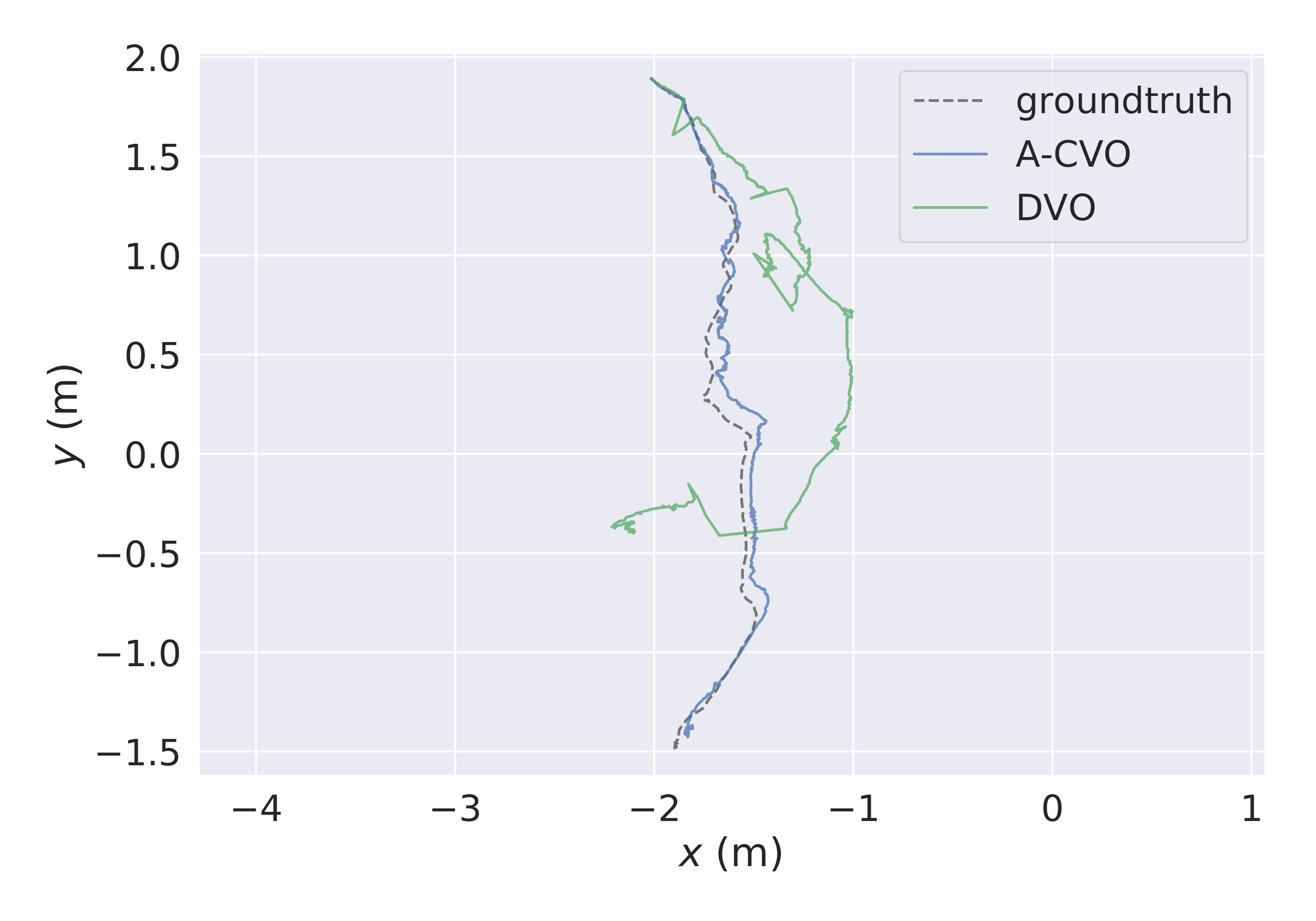}}\\ \vspace{-7mm}
    \subfloat{\includegraphics[page=2,width=\columnwidth]{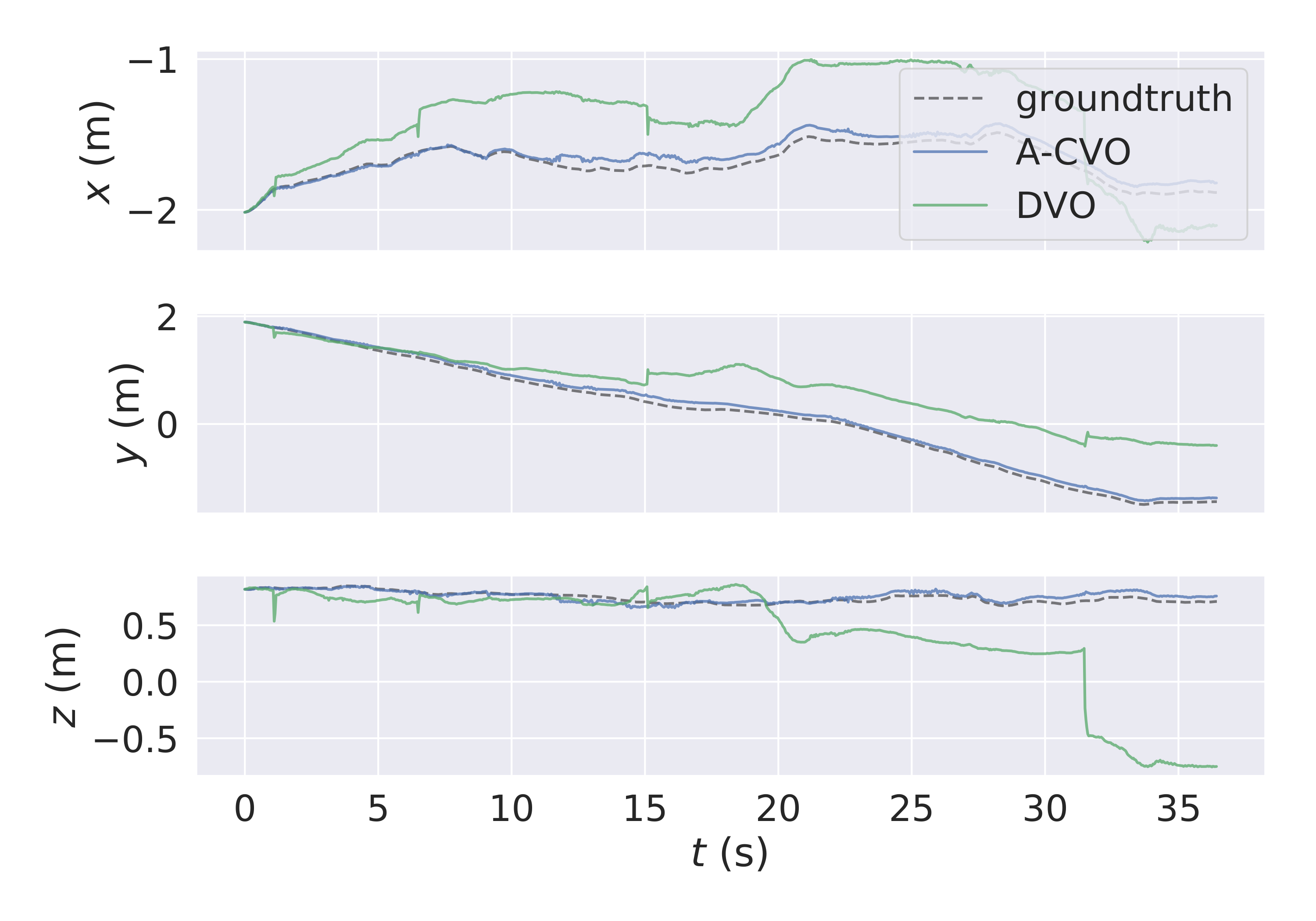}} \vspace{-2mm}
    \caption{The top figure shows trajectories of the proposed adaptive continuous visual odometry (A-CVO), dense visual odometry (DVO)~\cite{kerl2013robust}, and ground truth for \texttt{fr3/structure_notexture_near} sequence of the RGB-D benchmark in~\cite{sturm12iros}. The bottom figures show the x, y, and z trajectories vs. time. In the absence of texture, A-CVO performs well and almost follows the ground truth trajectory.}
    \label{fig:first_fig}
    \vspace{-5mm}
\end{figure}

Robust visual tracking has become a core aspect of state-of-the-art robotic perception and navigation in both structured and unstructured indoor and outdoor~\cite{klose2013efficient,kerl2013dense,kerl2013robust,engel2014lsd,engel2018direct,vidal2018ultimate,bryner2019event}. Hence, this work contributes to the foundations of robotic perception and autonomous systems via a continuous sensor registration framework enhanced by an adaptive hyperparameter learning strategy. In particular, this work has the following contributions:
\begin{enumerate}
    \item We extend the continuous visual odometry framework for RGB-D cameras to an adaptive framework via online hyperparameter learning. We also perform a sensitivity analysis of the problem and propose a systematic way to choose the sparsification threshold discussed in~\cite{MGhaffari-RSS-19}.
    \item We generalize the appearance (color) information inner product in~\cite{MGhaffari-RSS-19} to a kernelized form that improves the performance. With this improvement alone, the experimental evaluations show that the original continuous visual odometry is intrinsically robust and its performance is similar to that of the state-of-the-art robust dense (and direct) RGB-D visual odometry method~\cite{kerl2013robust}.
    \item We evaluate the proposed algorithm using the publicly available RGB-D benchmark in~\cite{sturm12iros} and make the software for the developed algorithm publicly available\footnote{Software is available for download at \href{https://github.com/MaaniGhaffari/cvo-rgbd}{\url{https://github.com/MaaniGhaffari/cvo-rgbd}}}. 
\end{enumerate}

The remainder of this paper is organized as follows. The problem setup is given in \S\ref{sec:problem}. The adaptive continuous visual odometry framework is discussed in \S\ref{sec:acvo}. The sensitivity analysis of the problem is provided in \S\ref{sec:sensitivity}. The experimental results are presented in \S\ref{sec:exp}. Finally, \S\ref{sec:conclusion} concludes the paper and discusses future research directions.

\section{Problem Setup}
\label{sec:problem}

Consider two (finite) collections of points, $X=\{x_i\}$, $Z=\{z_j\}\subset \mathbb{R}^3$. We want to determine which element \mbox{$h\in \mathrm{SE}(3)$}, where \mbox{$R \in \mathrm{SO}(3)$} and \mbox{$T \in \mathbb{R}^3$}, aligns the two point clouds $X$ and $hZ = \{hz_j\}$ the ``best.'' To assist with this, we will assume that each point contains information described by a point in an inner product space, $(\mathcal{I},\langle\cdot,\cdot\rangle_{\mathcal{I}})$. To this end, we will introduce two labeling functions, $\ell_X:X\to\Ical$ and $\ell_Z:Z\to\Ical$.

In order to measure their alignment, we will be turning the clouds, $X$ and $Z$, into functions $f_X,f_Z:\mathbb{R}^3\to\Ical$ that live in some reproducing kernel Hilbert space, $(\Hcal,\langle\cdot,\cdot\rangle_{\mathcal{H}})$. The action, $\mathrm{SE}(3) \curvearrowright \mathbb{R}^3$ induces an action $\mathrm{SE}(3) \curvearrowright C^\infty(\mathbb{R}^3)$ by $h.f(x) := f(h^{-1}x)$. Inspired by this observation, we will set $h.f_Z := f_{h^{-1}Z}$.

\begin{problem}\label{prob:problem}
	The problem of aligning the point clouds can now be rephrased as maximizing the scalar products of $f_X$ and $h.f_Z$, i.e., we want to solve
	\begin{equation}\label{eq:max}
		\argmax_{h\in \mathrm{SE}(3)} \, F(h),\quad F(h):= \langle f_X, h.f_Z\rangle_{\mathcal{H}}.
	\end{equation}
\end{problem}
\subsection{Constructing the functions}

We follow the same steps in~\cite{MGhaffari-RSS-19} with an additional step in which we use the kernel trick to kernelize the information inner product. For the kernel of our RKHS, $\Hcal$, we first choose the squared exponential kernel $k:\mathbb{R}^3\times\mathbb{R}^3\to\mathbb{R}$:
\begin{equation}\label{eq:k}
k(x,z) = \sigma^2\exp\left(\frac{-\lVert x-z\rVert_3^2}{2\ell^2}\right),
\end{equation}
for some fixed real parameters (hyperparameters) $\sigma$ and $\ell$, and $\lVert\cdot\rVert_3$ is the standard Euclidean norm on $\mathbb{R}^3$. This allows us to turn the point clouds to functions via
\begin{align}
	\nonumber f_X(\cdot) &:= \sum_{x_i\in X} \, \ell_X(x_i) k(\cdot,x_i), \\
	f_Z(\cdot) &:= \sum_{z_j\in Z} \, \ell_Z(z_j) k(\cdot,z_j).
\end{align}
We can now define the inner product of $f_X$ and $f_Z$ by
\begin{equation}\label{eq:scalar}
\langle f_X,f_Z\rangle_{\Hcal} := \sum_{\substack{x_i\in X, z_j\in Z}} \, \langle \ell_X(x_i),\ell_Z(z_j)\rangle_{\mathcal{I}} \cdot k(x_i,z_j).
\end{equation}

We use the well-known kernel trick in machine learning~\cite{bishop2006pattern,rasmussen2006gaussian,murphy2012machine} to substitute the inner products in~\eqref{eq:scalar} with the appearance (color) kernel. The kernel trick can be applied to carry out computations implicitly in the high dimensional space, which leads to computational savings when the dimensionality of the feature space is large compared to the number of data points~\cite{rasmussen2006gaussian}. After applying the kernel trick to~\eqref{eq:scalar}, we get
\begin{align}
\label{eq:newscalar}
    \nonumber \langle f_X,f_Z\rangle_{\Hcal} =& \sum_{\substack{x_i\in X, z_j\in Z}} \,  k_c(\ell_X(x_i),\ell_Z(z_j)) \cdot k(x_i,z_j) \\
    :=& \sum_{\substack{x_i\in X, z_j\in Z}} \,  c_{ij} \cdot k(x_i,z_j),
\end{align}
where we choose $k_c$ to be also the squared exponential kernel with fixed real hyperparameters $\sigma_c$ and $\ell_c$ that are set independently.

\section{Adaptive Continuous Visual Odometry via Online Hyperparameter Learning}
\label{sec:acvo}

The length-scale of the kernel, $\ell$, is an important hyperparameter that affects the performance and convergence of the algorithm significantly. In the original framework in~\cite{MGhaffari-RSS-19}, $\ell$ was set using a fixed set of conditions within the solver to reduce the length-scale as the algorithm reached a local minimum. Intuitively, large values of $\ell$ encourage higher correlations between points that are far apart from each other; and small values of $\ell$ encourage the algorithm to focus on only points that are very close to each other with respect to the distance metric of the kernel (here we use the Euclidean distance). This latter case results in faster convergence and it can be thought of as refinement steps where the target and source clouds are already almost aligned.

Now the question to answer is how can we tune $\ell$ automatically and online at each iteration so that the overall registration performance is maximized? In this section, we provide a solution that is based on a greedy gradient descent search. As we will see, this approach is highly appealing due to its simplicity and the gain in performance. We first revisit Problem~\ref{prob:problem}. The maximization of the inner product is a reduced form of the original cost $J(h) := \lVert f_X - h.f_Z \rVert^2_{\Hcal}$ and the fact that $\mathrm{SE}(3) \curvearrowright C^\infty(\mathbb{R}^3)$ is an isometry. That is
\begin{align}
\label{eq:cost}
    \nonumber \lVert f_X &- h.f_Z \rVert^2_{\Hcal} = \lVert f_X \rVert^2_{\Hcal} + \lVert f_Z \rVert^2_{\Hcal} - 2 \langle f_X, h.f_Z\rangle_{\mathcal{H}} \\
    \nonumber &= \sum_{\substack{x_i,x_j\in X}} \,  \alpha_{ij} \cdot k(x_i,x_j) + \sum_{\substack{z_i, z_j\in Z}} \,  \beta_{ij} \cdot k(z_i,z_j) \\
    &- 2 \sum_{\substack{x_i\in X, z_j\in Z}} \,  c_{ij} \cdot k(x_i,h^{-1}z_j),
\end{align}
where coefficients $\alpha_{ij}$ and $\beta_{ij}$ are defined for each function's inner product with itself similar to~\eqref{eq:newscalar}.

Computing the gradient of~\eqref{eq:cost} with respect to $\ell$ is straightforward and is given by
\begin{align}
    \label{eq:ellgrad}
        \nonumber \frac{\partial J(\ell)}{\partial \ell} &= \frac{\partial J}{\partial k} \cdot \frac{\partial k}{\partial \ell} \\
        \nonumber &= \frac{1}{\ell^3} \Biggl[\ \sum_{\substack{x_i,x_j\in X}} \,  \alpha_{ij} \cdot p_{ij} \cdot k(x_i,x_j) \\ 
        \nonumber &+ \sum_{\substack{z_i,z_j\in Z}} \,  \beta_{ij} \cdot q_{ij} \cdot k(z_i,z_j) \\ 
        &- 2 \sum_{\substack{x_i\in X, z_j\in Z}} \,  c_{ij} \cdot r_{ij} \cdot k(x_i,h^{-1}z_j) \Biggr],
\end{align}
where we defined $p_{ij} := \lVert x_i-x_j\rVert_3^2$, $q_{ij} := \lVert z_i-z_j\rVert_3^2$, and $r_{ij} := \lVert x_i-z_j\rVert_3^2$. Then using the following update (integration) rule we find the length-scale for the next iteration,
\begin{equation}
    \label{eq:ellupdate}
    \ell_{k+1} = \ell_k - \gamma_{\ell} \cdot \frac{\partial J(\ell_k)}{\partial \ell_k},
\end{equation}
where $\gamma_{\ell}$ is the step size (learning rate).

This strategy alone can lead to failure or extremely poor performance based on our observations. The reason is that CVO uses semi-dense data and in the absence of structure or texture in the environment the gradient can be weak or not well-behaved. To address this problem, we can simply define a search interval for the length-scale as $\ell \in [\ell_{min}, \ell_{max}]$. This additional step not only keeps $\ell$ in a feasible region but also allows the algorithm to detect when tracking is difficult and issue a warning message. To improve the convergence, when $\ell \geq \ell_{max}$, we reduce both $\ell$ and $\ell_{max}$ by a reduction factor, $0 < \lambda_{\ell} < 1$, and continue as before. 


\section{Sensitivity Analysis}
\label{sec:sensitivity}
Understanding how $k$ in equation \eqref{eq:k} depends on $\ell$ is a surprisingly delicate problem which, surprisingly, offers a systematic way to choose the kernel sparsification threshold (see Table \ref{tab:parameters}). Consider the following normalization of $k$:
\begin{equation}
g(s) = \exp\left(-\frac{1}{2s^2}\right),\quad s = \frac{\ell}{\lVert x-y\rVert},
\end{equation}
so $\sigma^2g(s) = k(x,y)$. Suppose we want to find an approximation of $g$ as $s$ gets small; this is equivalent to understanding $k(x,y)$ as $\lVert x-y\rVert >> \ell$. Performing a Taylor expansion of $g(s)$ about $s=0$ results in the zero function. A simple enough calculation shows that
\begin{equation}
    \frac{d^k}{ds^k} g(s) = p_k(s)g(s),
\end{equation}
where $p_k$ is some rational function. Due to the fact that $g$ is exponential, we have that
\begin{equation}
    \lim_{s\to 0^+} Q(s)g(s) = 0,
\end{equation}
for \textit{any} rational function $Q$. This shows that the Taylor series of $g$ about $s=0$ is trivially zero. (The underlying reason for the Taylor series being zero while the function is not is because $g$ is not \textit{analytic} at $s=0$. In fact, if we view $g$ as a complex function there is an \textit{essential singularity} at $s=0$, see \S 5.6 in \cite{saff2003fundamentals}.)

\begin{figure}[t]
    \centering
    \includegraphics[width=\columnwidth]{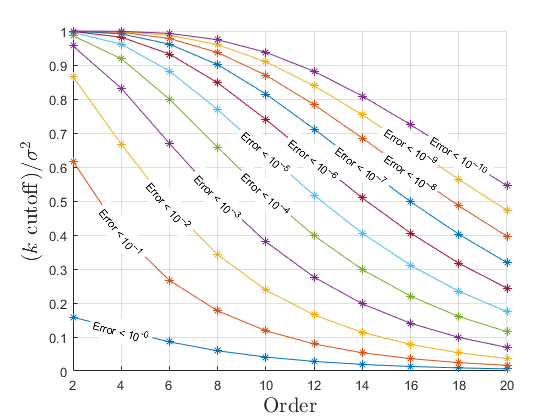}
    \caption{This shows the cutoff values for $k$ based on the order of the approximation and the required error tolerances. For example, if we use a $6^{th}$-order expansion and we require an error of less than $10^{-3}$, then all points where $k(x,y) < 0.6694\sigma^2$ need to be ignored.}
    \label{fig:cut_offs}
\end{figure}

Rather than expanding about $s=0$ (where points are far apart), we can expand about $s=\infty$ (where points are close together). This results in the following expansion:
\begin{equation}\label{eq:laurent}
    g(s) \sim 1 - \frac{1}{2}s^{-2} + \frac{1}{8}s^{-4} - \frac{1}{48}s^{-6} + \ldots.
\end{equation}
While this approximation is accurate when $s$ is large, as $s$ approaches zero this approximation falls apart. The exact function, $g$, approaches zero as $s\to 0$ but the approximation has a pole at zero regardless of order. This motivates a minimum cutoff for $s$ such that \eqref{eq:laurent} has a well controlled error. By applying this cutoff to the original function $k$, we can obtain a kernel sparsification threshold that guarantees error bounds in the approximation \eqref{eq:laurent}. A plot of these values is shown in Fig. \ref{fig:cut_offs}.

\section{Experimental Results}
\label{sec:exp}
We now present experimental evaluations of the proposed method Adaptive CVO (A-CVO). We compare A-CVO with the original CVO~\cite{MGhaffari-RSS-19} and the state-of-the-art direct (and dense) RGB-D visual odometry (DVO)~\cite{kerl2013dense}. Since the original DVO source code requires outdated ROS dependency~\cite{Kerl2013DVOrepo}, we reproduced DVO results using the version provided by Matthieu Pizenberg~\cite{MatthieuP2019DVO}, which only removes the dependency for ROS while maintains the DVO core source code unchanged. We also include the DVO results of Kerl et al.~\cite{kerl2013dense} for reference. We refer to the reproduced DVO results as \emph{DVO} and the results directly taken from~\cite{kerl2013dense} as \mbox{\emph{Kerl et al.\cite{kerl2013dense}}}.

\begin{table}[t]
\centering
\caption{ Parameters used for evaluation using TUM RGB-D Benchmark, similar values are chosen for all experiments. The kernel characteristic length-scale is chosen to be adaptive as the algorithm converges; intuitively, we prefer a large neighborhood of correlation for each point, but as the algorithm reaches the convergence reducing the local correlation neighborhood allows for faster convergence and better refinement.}
\resizebox{\columnwidth}{!}{
\footnotesize
\begin{tabular}{lcr}
\toprule
    Parameters & Symbol & Value \\
        \midrule
        Transformation convergence threshold & $\epsilon_1$ & $1\mathrm{e}{-5}$ \\
        Gradient norm convergence threshold & $\epsilon_2$ & $5\mathrm{e}{-5}$ \\
        Minimum step length & & $0.2$ \\
        Kernel sparsification threshold & & $8.315\mathrm{e}{-3}$ \\
        Spatial kernel initial length-scale & $\ell_{init}$ & $0.1$ \\
        Spatial kernel signal variance & $\sigma$ & $0.1$  \\
        Spatial kernel minimum length-scale (A-CVO) & $\ell_{min}$ & $0.039$ \\
        Spatial kernel maximum length-scale (A-CVO) & $\ell_{max}$ & $0.15$ \\
        Color kernel length-scale & $\ell_c$ & $0.1$ \\
        Color kernel signal variance & $\sigma_c$ & $1$  \\
        $\ell$ integration step size (A-CVO) & $\gamma_{\ell}$ & $0.3$\\
        $\ell$ reduction factor (A-CVO) & $\lambda_{\ell}$ & $0.7$ \\
\bottomrule
\end{tabular}
}
\label{tab:parameters}
\squeezeup\squeezeup
\end{table}

\begin{figure*}
    \centering
    \subfloat{\includegraphics[width=0.33\textwidth]{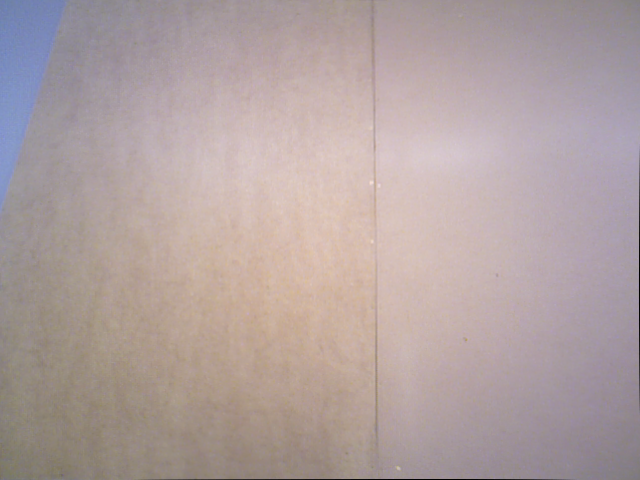}} 
    \hfill
    \subfloat{\includegraphics[width=0.33\textwidth]{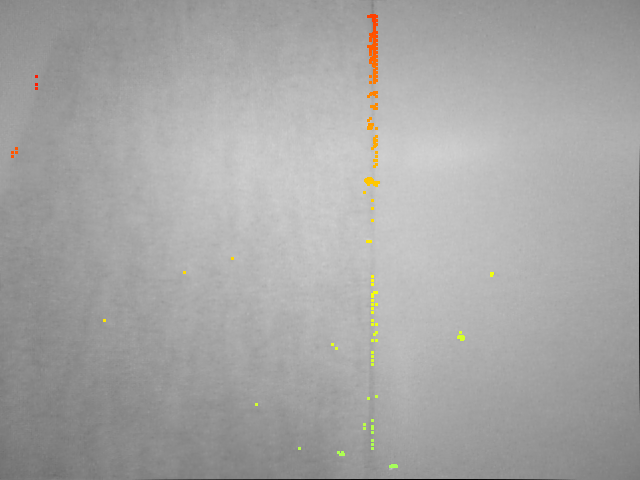}}
    \hfill
    \subfloat{\includegraphics[width=0.33\textwidth]{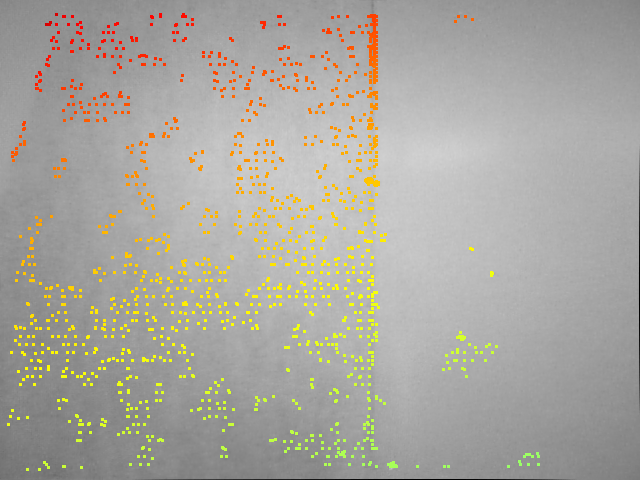}}\\ \vspace{-2mm}
    
    \subfloat{\includegraphics[width=0.33\textwidth]{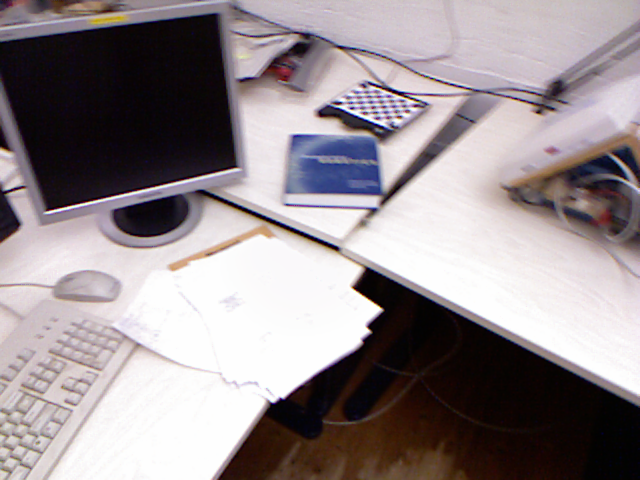}}
    \hfill
    \subfloat{\includegraphics[width=0.33\textwidth]{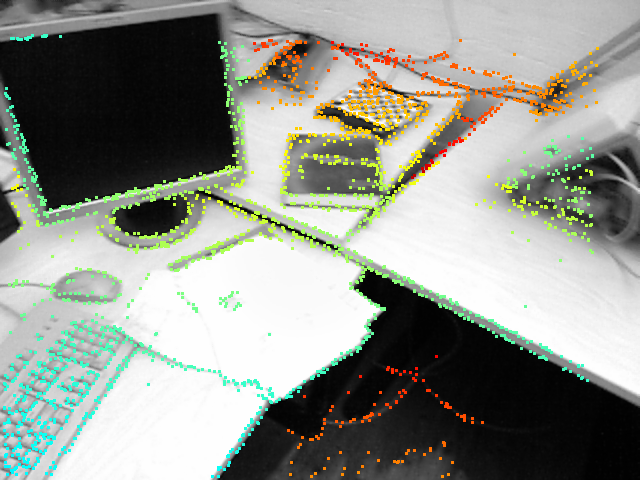}}
    \hfill
    \subfloat{\includegraphics[width=0.33\textwidth]{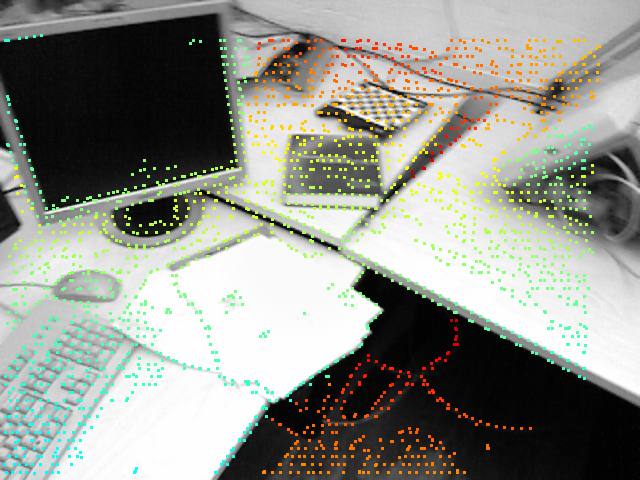}}
    \caption{The visualization of the point selection mechanism adopted from DSO~\cite{engel2018direct}. The top row shows image \texttt{1341840842.006342.png} in \texttt{fr3/nostructure_notexture_far} sequence. The bottom row shows image \texttt{1305031453.359684.png} in \texttt{fr1/desk} sequence. The leftmost images are the original image recorded by a Microsoft Kinect. The images in the middle show the points selected by the DSO point selection algorithm. The top right image shows the points selected when the number of points selected by DSO is insufficient. Extra points are selected by downsampling the highlighted edge computed using the Canny edge detector~\cite{opencv_library}. The bottom right image shows the points selected solely by downsampling of the Canny edge detector output. In this case, since the DSO point selector already picked up enough points, the points from the Canny detector are not used. The points highlighted in the images are made 9 times bigger (i.e. 9 pixels are drawn) to make the visualization more clear.} 
    \label{fig:point_selection} 
    \squeezeup
\end{figure*}

\begin{table*}[t]
    \centering
    \caption{The RMSE of Relative Pose Error (RPE) for \texttt{fr1} sequences. The trans. columns show the RMSE of the translational drift in $\mathrm{m}/\sec$ and the rot. columns show the RMSE of the rotational error in $\mathrm{deg}/\sec$. The Average* shows the average result excluding \texttt{fr1/floor} sequence since Kerl et al.\cite{kerl2013dense} reported a failure on that sequence. The rotational errors for Kerl et al. were left empty for they were not reported in the original paper~\cite{kerl2013dense}. There's no corresponding validation datasets for \texttt{fr1/teddy} and \texttt{fr1/floor}. The results show that A-CVO out-performs the other two methods on both training and validation sets of \texttt{fr1}.}
    \resizebox{\textwidth}{!}{
    \begin{tabular}{lcccccccc|cccccccc}
        \toprule
         & \multicolumn{8}{c}{Training} & \multicolumn{8}{c}{Validation} \\
        \midrule
          & \multicolumn{2}{c}{CVO~\cite{MGhaffari-RSS-19}} & \multicolumn{2}{c}{A-CVO} & \multicolumn{2}{c}{Kerl et al.\cite{kerl2013dense}} & \multicolumn{2}{c}{DVO} & \multicolumn{2}{c}{CVO~\cite{MGhaffari-RSS-19}} & \multicolumn{2}{c}{A-CVO} & \multicolumn{2}{c}{Kerl et al.\cite{kerl2013dense}} & \multicolumn{2}{c}{DVO} \\
         Sequence & Trans. & Rot. & Trans. & Rot. & Trans. & Rot. & Trans. & Rot. & Trans. & Rot. & Trans. & Rot. & Trans. & Rot. & Trans. & Rot. \\
        \midrule
        fr1/desk    & 0.0486 & 2.4860 & 0.0375 & \bf 2.1456 & \bf 0.0360 & n/a & 0.0387 & 2.3589 &
                      0.0401 & 2.0148 & 0.0431 & \bf 1.8831 & \bf 0.0350 & n/a & 0.0371 & 2.0645  \\
        fr1/desk2   & 0.0535 & 3.0383 & \bf 0.0489 & \bf 2.5857 & 0.0490 & n/a & 0.0583 & 3.6529 &
                      0.0225 & 1.7691 & 0.0224 & \bf 1.6584 & \bf 0.0200 & n/a & 0.0208 & 1.7416  \\ 
        fr1/room    & 0.0560 & 2.4566 & 0.0529 & \bf 2.2750 & 0.0580 & n/a & \bf 0.0518 & 2.8686 &
                      \bf 0.0446 & \bf 3.9183 & 0.0465 & 3.9669 & 0.0760 & n/a & 0.2699 & 7.4144  \\
        fr1/360     & \bf 0.0991 & 3.0025 & 0.0993 & \bf 3.0125 & 0.1190 & n/a & 0.1602 & 4.4407 &
                      0.1420 & 3.0746 & 0.0995 & \bf 2.2177 & \bf 0.0970 & n/a & 0.2811 & 7.0876  \\
        fr1/teddy   & 0.0671 & 4.8089 & \bf 0.0553 & \bf 2.2342 & 0.0600 & n/a & 0.0948 & 2.5495 & 
                      n/a    & n/a    & n/a    & n/a    & n/a    & n/a & n/a    & n/a  \\
        fr1/floor   & 0.0825 & 2.3745 & 0.0899 & 2.2904 & fail   & n/a & \bf 0.0635 & \bf 2.2805 & 
                      n/a    & n/a    & n/a    & n/a    & n/a    & n/a & n/a    & n/a  \\
        fr1/xyz     & 0.0240 & 1.1703 & \bf 0.0236 & \bf 1.1682 & 0.0260 & n/a & 0.0327 & 1.8751 &
                      0.0154 & 1.3872 & \bf 0.0150 & \bf 1.2561 & 0.0470 & n/a & 0.0453 & 3.0061 \\
        fr1/rpy     & 0.0457 & 3.3073 & 0.0425 & 3.0497 & 0.0400 & n/a & \bf 0.0336 & \bf 2.6701 &
                      0.1138 & 3.6423 & \bf 0.0799 & \bf 2.4335 & 0.1030 & n/a & 0.3607 & 7.9991  \\
        fr1/plant   & 0.0316 & 1.9973 & 0.0347 & 1.8580 & 0.0360 & n/a & \bf 0.0272 & \bf 1.5523 &
                      0.0630 & 4.9185 & \bf 0.0591 & 4.1925 & 0.0630 & n/a & 0.0660 & \bf 2.5865  \\
        \midrule
        Average*    & 0.0532 & 2.7834 & \bf 0.0493 & \bf 2.2911 & 0.0530 & n/a & 0.0622 & 2.7460 &
                      - & - & - & - & - & n/a & - & - \\
        Average all & 0.0561 & 2.7380 & \bf 0.0534 & \bf 2.2910 & n/a & n/a & 0.0623 & 2.6943 &
                      0.0631 & 2.9607 & \bf 0.0522 & \bf 2.5155 & 0.0630 & n/a & 0.1544 & 4.5571 \\
        \bottomrule
    \end{tabular}
    }
    \label{tab:fr1_results}
\end{table*}

\subsection{Experimental Setup}

To improve the computational efficiency, we adopted a similar approach to Direct Sparse Visual Odometry (DSO) by Engel et al.~\cite{engel2014lsd} to create a semi-dense point cloud (around 3000 points) for each scan. To prevent insufficient points being selected in environments that lack rich visual information, we also used a Canny edge detector~\cite{canny1987computational} from OpenCV~\cite{opencv_library}. When the points selected by the DSO point selector are less than one-third of the desired number of points, more points will be selected by downsampling the pixels highlighted by the Canny detector. While generating the point cloud, RGB values are first transformed into HSV colormap and normalized. The normalized HSV values are then combined with the normalized intensity gradients and utilized as the labels of the selected points in the color space. For all experiments, we used the same set of parameters, which are listed in Table~\ref{tab:parameters}.

All experiments are performed on a Dell XPS15 9750 laptop with Intel i7-8750H CPU (6 cores with 2.20 GHz each) and 32GB RAM. The source code is implemented in C++ and compiled with the Intel Compiler. The kernel computations are parallelized using the Intel Threading Building Blocks (TBB)~\cite{intel2019tbb}. Using compiler auto-vectorization and the parallelization, the average time for frame-to-frame registration is 0.5 $\sec$. The frame-to-frame registration time for the original CVO is 0.2 $\sec$ (5 $\mathrm{Hz}$).

\begin{figure*}[th]
    \centering
    \subfloat{\includegraphics[trim={1cm 1cm 0.5cm 0},clip,width=0.333\textwidth]{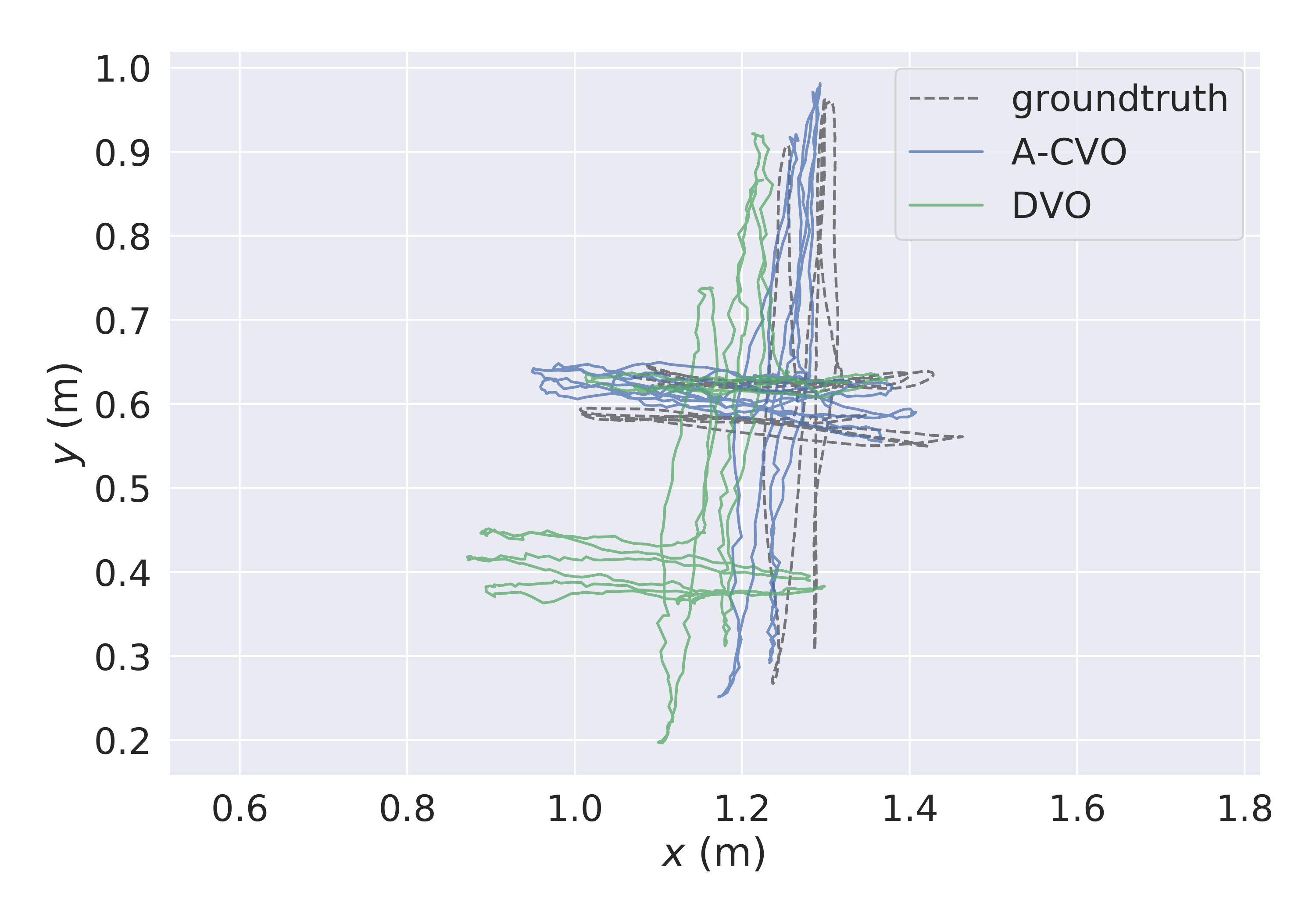}}
    \subfloat{\includegraphics[trim={1cm 1cm 0.5cm 0},clip,page=2,width=0.333\textwidth]{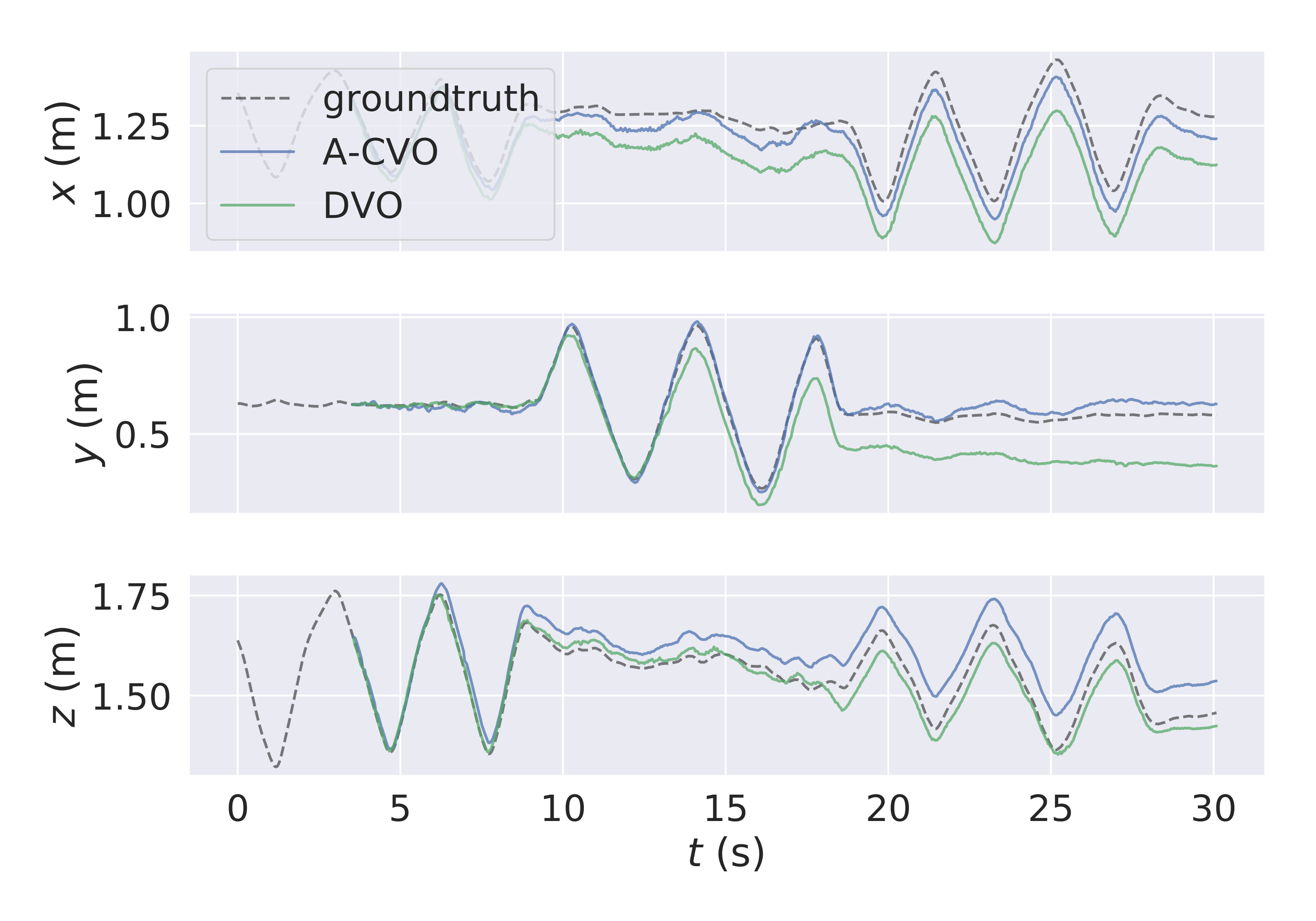}} 
    \subfloat{\includegraphics[trim={1cm 1cm 0.5cm 0},clip,page=3,width=0.333\textwidth]{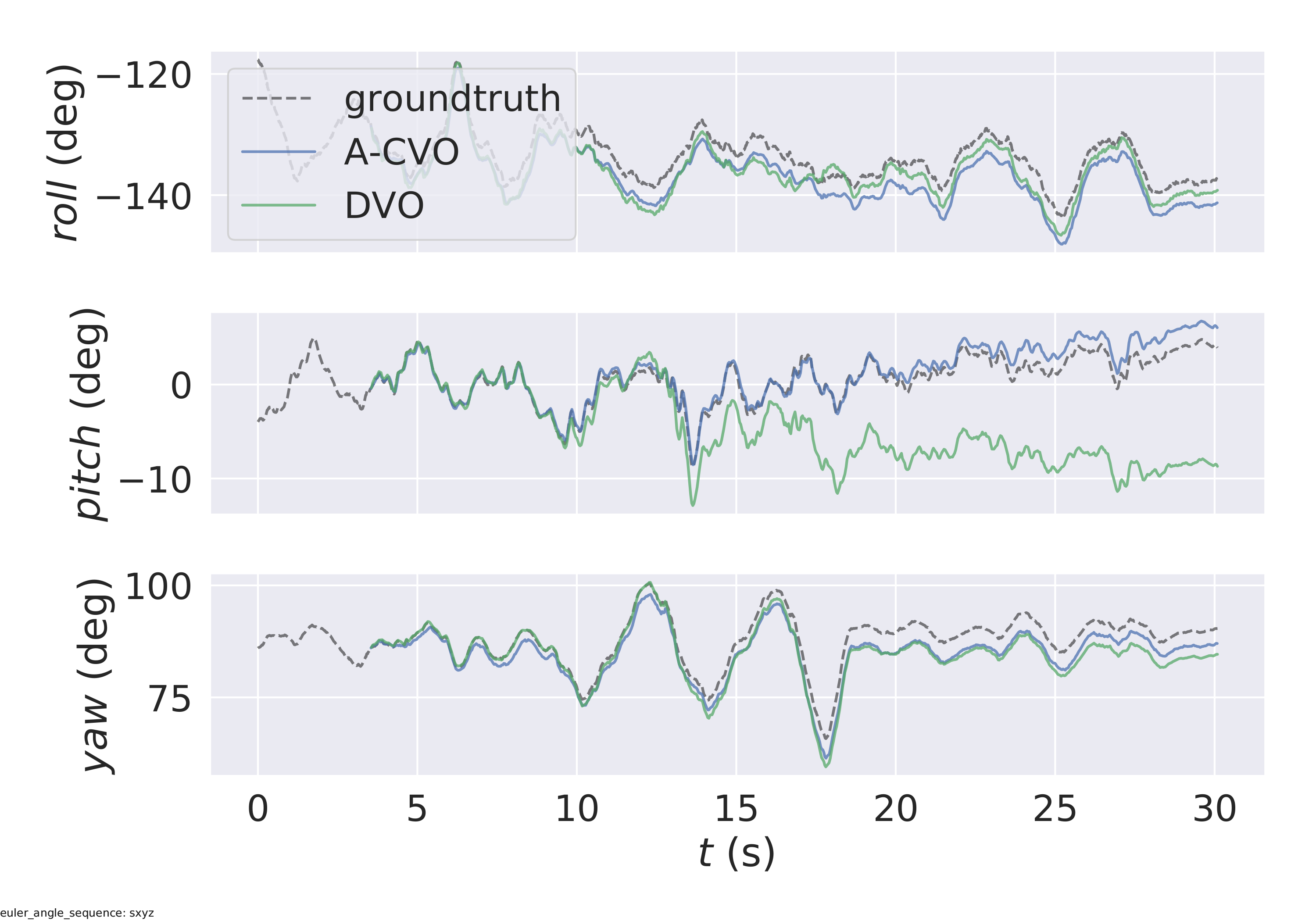}} 
    \caption{The trajectory comparison of A-CVO, DVO, and groundtruth for \texttt{fr1/xyz} sequence. The left figure shows the trajectory in the xy plane. The middle figure shows the x, y, z trajectory with respect to time. (Note that it's not the error plot.) The figure in the right shows the angles of roll, pitch, and yaw with respect to time. This sequence contains a repetitive motion and can show the repeatability of a method. As shown in the figures, A-CVO can follow the groundtruth trajectory while DVO drifts in the y direction.}
    \label{fig:fr1xyz}
    \squeezeup
\end{figure*}

\subsection{TUM RGB-D Benchmark}

We performed experiments on two parts of \mbox{RGB-D} SLAM dataset and benchmark by the Technical University of Munich~\cite{sturm12iros}. This dataset was collected indoors with a Microsoft Kinect using a motion capture system as a proxy for ground truth trajectory. For all tracking experiments, the entire images were used sequentially without any skipping, i.e., at full frame rate. We evaluated A-CVO, CVO, and DVO on the training and validation sets for all the fr1 sequences and the structure versus texture sequences. \mbox{RGB-D} benchmark tools~\cite{sturm12iros} were then used to evaluate the Relative Pose Error (RPE) of all 3 methods, and evo~\cite{grupp2017evo} was utilized to visualize the trajectory. 

Table~\ref{tab:fr1_results} shows the Root-Mean-Squared Error (RMSE) of the RPE for \texttt{fr1} sequences. The Trans. columns show the RMSE of the translational drift in $\mathrm{m}/\sec$ and the Rot. columns show the RMSE of the rotational drift in $\mathrm{deg}/\sec$. The Average* shows the average result by excluding \texttt{fr1/floor} sequence since Kerl et. al reported failure on that sequence~\cite{kerl2013dense}. The rotational errors were not reported in the original paper~\cite{kerl2013dense}. There are no corresponding validation sequences for \texttt{fr1/teddy} and \texttt{fr1/floor}. A-CVO improves the performance over CVO and outperforms DVO on both translational and rotational metrics. On the training sequences, A-CVO reduces the average translational error of CVO by 7.2$\%$, and on the validation sequences, the improvement reaches to 17.2$\%$. A-CVO has a 6.9$\%$ lower translational error than Kerl et al. on the training set (excluding the failure case). On the validation set, A-CVO has 17.1$\%$ improved performance compared with Kerl et al. which shows A-CVO can generalize across different scenarios better. It is worth noting that CVO is intrinsically robust and its performance is similar to that of the state-of-the-art robust dense (and direct) RGB-D visual odometry method~\cite{kerl2013robust}. Next experiment will further reveal that CVO has the advantage of performing well in extreme environments that lack rich structure or texture.

\begin{table*}[th!]
    \centering
    \caption{The RMSE of Relative Pose Error (RPE) for the structure v.s texture sequence. The Trans. columns show the RMSE of the translational drift in $\mathrm{m}/\sec$ and the Rot. columns show the RMSE of the rotational error in $\mathrm{deg}/\sec$. The Average* shows the average value excluding \texttt{fr3/nostructue_notexture_near} and \texttt{fr3/nostructure_notexture_far}. The \checkmark means the sequence has structure/texture and $\times$ means the sequence does not have structure/texture. The results show while DVO performs better in structure and texture case, A-CVO has significantly better accuracy in the environments that lack structure and texture.}
    \resizebox{\textwidth}{!}{
    \begin{tabular}{lllcccccccc|cccccccc}
        \toprule
         & & & \multicolumn{8}{c}{Training} & \multicolumn{8}{c}{Validation} \\
        \midrule
           \multicolumn{3}{c}{Sequence}  & \multicolumn{2}{c}{CVO~\cite{MGhaffari-RSS-19}} & \multicolumn{2}{c}{A-CVO} & \multicolumn{2}{c}{Kerl et al.\cite{kerl2013dense}} & \multicolumn{2}{c}{DVO} & \multicolumn{2}{c}{CVO~\cite{MGhaffari-RSS-19}} & \multicolumn{2}{c}{A-CVO} & \multicolumn{2}{c}{Kerl et al.\cite{kerl2013dense}} & \multicolumn{2}{c}{DVO} \\
         \multicolumn{3}{l}{structure-texture-dist.} & Trans. & Rot. & Trans. & Rot. & Trans. & Rot. & Trans. & Rot. & Trans. & Rot. & Trans. & Rot. & Trans. & Rot. & Trans. & Rot. \\
        \midrule
        $\times$   & \checkmark & near & 0.0279 & 1.3470 & \bf 0.0267 & \bf 1.3033 &  0.0275 & n/a &  0.0563 & 1.7560 &
                                         \bf0.0310 & 1.6367 & 0.0313 & 1.6089 & n/a & n/a & 0.0315 & \bf1.1498  \\
        $\times$   & \checkmark & far  & \bf 0.0609 & 1.2342 & 0.0613 & \bf 1.1985 &  0.0730 & n/a &  0.1612 & 3.4135 & 
                                         0.1374 & 2.3929 & \bf0.1158 & \bf2.0423 & n/a & n/a & 0.5351 & 8.2529 \\
        \checkmark & $\times$   & near & 0.0221 & \bf 1.3689 & 0.0261 & 1.5059 &  \bf 0.0207 & n/a &  0.1906 & 10.6424 & 
                                         0.0465 & 2.0359 & \bf0.0405 & \bf1.7665 & n/a & n/a & 0.1449 & 4.9022 \\
        \checkmark & $\times$   & far  & 0.0372 & 1.3061 & \bf 0.0323 & \bf 1.1114 &  0.0388 & n/a &  0.1171 & 2.4044 & 
                                         0.0603 & 1.8142 & \bf0.0465 & \bf1.3874 & n/a & n/a & 0.1375 & 2.2728 \\
        \checkmark & \checkmark & near & 0.0236 & 1.2972 & 0.0367 & 1.6223 & 0.0407 & n/a & \bf 0.0175 & \bf 0.9315 & 
                                         0.0306 & 1.8694 & 0.0394 & 2.2864 & n/a & n/a & \bf0.0217 & \bf1.2653 \\
        \checkmark & \checkmark & far  & 0.0409 & 1.1640 & 0.0369 & 1.0236 &  0.0390 & n/a & \bf 0.0171 & \bf 0.5717 & 
                                         0.0616 & 1.4760 & 0.0446 & 1.1186 & n/a & n/a & \bf0.0230 & \bf0.6312 \\
        $\times$   & $\times$   & near & 0.2119 & 9.7944 & \bf 0.1790 & \bf 7.0098 &  n/a    & n/a & 0.3506 & 13.3127 &
                                         0.1729 & \bf5.8674 & \bf0.1568 & 6.8221 & n/a & n/a & 0.1747 & 6.0443 \\
        $\times$   & $\times$   & far  & \bf 0.0799 & \bf 3.0978 & 0.1151 & 3.8035 &  n/a    & n/a & 0.1983 & 6.8419 & 
                                         0.0899 & 2.6199 & \bf0.0805 & \bf2.4138 & n/a & n/a & 0.2000 & 6.5192 \\
        \midrule
		\multicolumn{3}{c}{Average*}    & \bf 0.0355 & \bf 1.2862 & 0.0367 & 1.2942 & 0.0400 & n/a & 0.0933 & 3.2866 & 
		                                 0.0612 & 1.8708 & \bf0.0530 & \bf1.7017 & n/a & n/a & 0.1490 & 3.0790 \\
		\multicolumn{3}{c}{Average all}    & \bf 0.0631 & 2.5762 & 0.0643 & \bf 2.3223 & n/a & n/a & 0.1386 & 4.9843 & 
		                                 0.0787 & 2.4640 & \bf0.0694 & \bf2.4307 & n/a & n/a & 0.1586 & 3.8797 \\
        \bottomrule
    \end{tabular}
    }
    \label{tab:fr3_results}
    \squeezeup
\end{table*}

\begin{figure*}[t]
    \centering
    \subfloat{\includegraphics[trim={1cm 1cm 0.5cm 0},clip,width=0.333\textwidth]{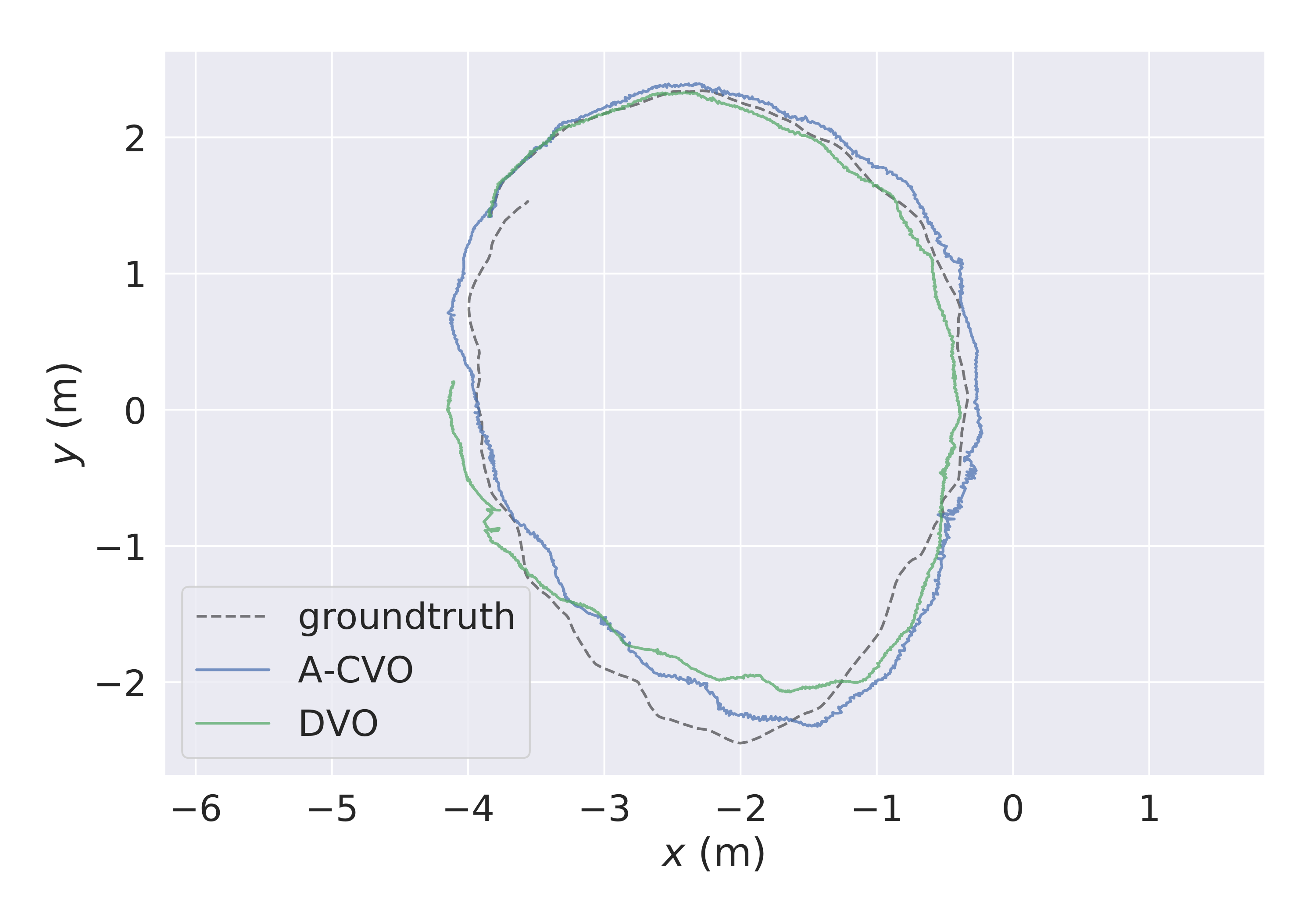}}
    \subfloat{\includegraphics[trim={1cm 1cm 0.5cm 0},clip,page=2,width=0.333\textwidth]{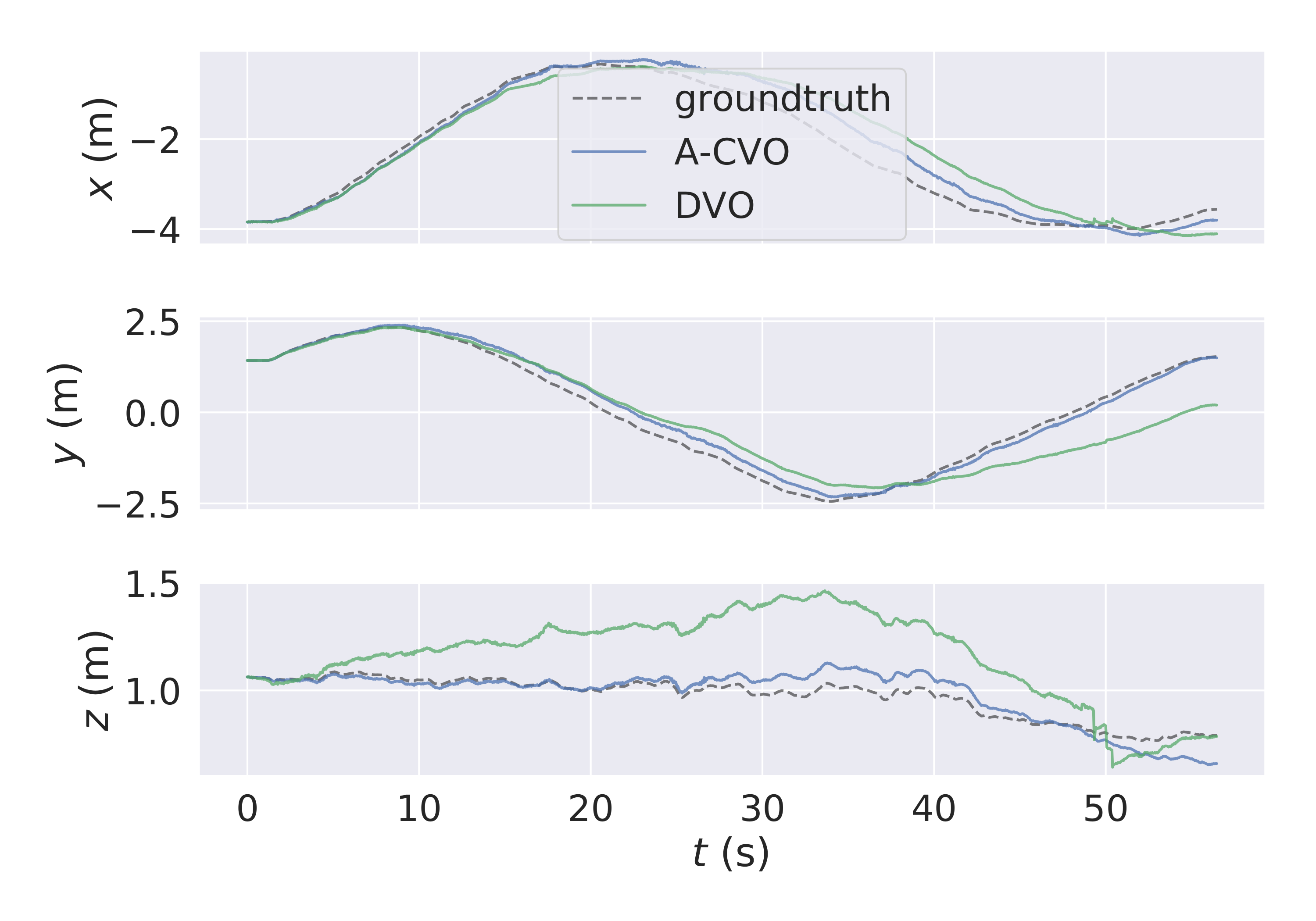}} 
    \subfloat{\includegraphics[trim={1cm 1cm 0.5cm 0},clip,page=3,width=0.333\textwidth]{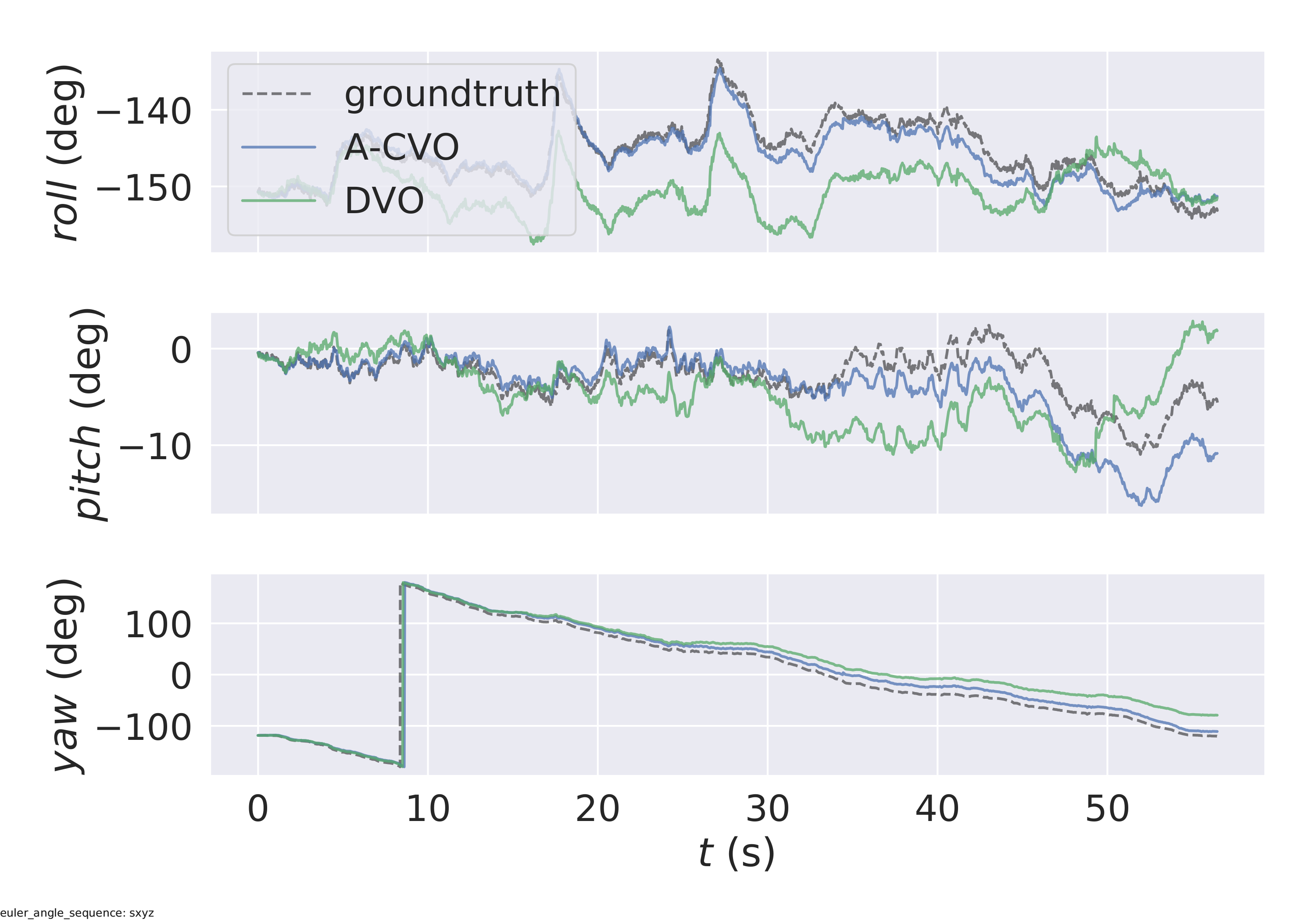}} 
    \caption{The comparison of A-CVO, DVO, and groundtruth trajectory for \texttt{fr3/nostructure_texture_near} sequence. The left figure shows the trajectory in the xy plane. The middle figure shows the x, y, z trajectory with respect to time. (Note that it's not the error plot.) The figure in the right shows the angles of roll, pitch, and yaw with respect to time. From the plot, we can see A-CVO follows the groundtruth trajectory better than DVO.}
    \label{fig:fr3something}
    \squeezeup
\end{figure*}

\subsection{Experiments using Structure vs. Texture Sequences}

Table~\ref{tab:fr3_results} shows the RMSE of RPE for the structure vs. texture sequences. This dataset contains image sequences in structure/nostructure and texture/notexture environments. As elaborated in~\cite{MGhaffari-RSS-19}, by treating point clouds as points in the function space (RKHS), CVO and A-CVO are inherently robust to the lack of features in the environment. A-CVO and CVO show the best performance on cases that either structure or texture is not rich in the environment. This reinforces the claim in~\cite{MGhaffari-RSS-19} that CVO is robust to such scenes.

However, by online hyperparameter learning, A-CVO allows the parameters to be adaptively varying with the environment without the need for manual tuning, which improves the performance over the original CVO. We also note that DVO has the best performance on the case where the environment contains rich texture and structure information. This can be because of two reasons: 1) CVO and A-CVO adopted a semi-dense point cloud construction from DSO~\cite{engel2018direct}, while DVO uses the entire dense image without subsampling. Although the semi-dense tracking approach of Engel et al.~\cite{engel2014lsd,engel2018direct} is computationally attractive and we advocate it, the semi-dense point cloud construction process used in this work is a heuristic process and might not necessarily capture the relevant information in each frame optimally; 2) DVO uses a motion prior as regularizer whereas CVO and A-CVO solely depend on the camera information with no regularizer. We conjecture this latter is the reason DVO, relative to the training set, does not perform well on validation sequences. The motion prior is a useful assumption when it is true! It can help to tune the method better on the training sets but if the assumption gets violated can lead to poor performance. The addition of an IMU sensor, of course, can improve the performance of all the compared methods and is an interesting future research direction.

\subsection{Discussions and Limitations}
\label{sec:discussions}



We have shown that A-CVO and CVO perform well across different indoor scenarios and different structure and texture conditions. The TUM RGB-D benchmark used in this paper was collected using a Microsoft Kinect for Xbox 360 which has a rolling shutter camera and is not designed for robotic applications. We observed that the blurred images due to camera motion are the most challenging frames for registration. The performance can degrade considerably as the extraction of the semi-dense structure cannot capture the structure and texture of the scene accurately. For example, a table edge that is usually a reliable part of an image, when blurred, can result in hallucinating multiple lines.  Although more recent cameras used in robotics often use global shutters, the problem is still relevant and should be addressed. Exploring point selection strategies to improve the performance on challenging frames is also an interesting topic as future work. 


The current implementation of CVO/A-CVO exploits vectorization and multi-threading which means the provided software gain additional performance benefits automatically as vector registers continue becoming wider. However, robotic applications require real-time software and more work is needed in order to achieve real-time performance using CPUs. An interesting research avenue to obtain real-time performance is a GPU implementation of the CVO/A-CVO.  


\section{Conclusion and Future Work}
\label{sec:conclusion}

We have developed an adaptive continuous visual odometry method for RGB-D cameras via online hyperparameter learning. The experimental results indicate that the original continuous visual odometry is intrinsically robust and its performance is similar to that of the state-of-the-art robust dense (and direct) RGB-D visual odometry method. Moreover, online learning of the kernel length-scale brings significant performance improvement and enables the method to perform better across different domains even in the absence of structure and texture in the environment.

In the future, we can use the invariant IMU model in~\cite{hartley2019contact} to predict the next camera pose and use the predicted pose as the initial guess in the A-CVO algorithm. This alone can increase the performance as the model performs an exact integration within a small time between two images. The integration of A-CVO into multisensor fusion systems~\cite{leutenegger2015keyframe,usenko2016direct,forster2017manifold,rhartley-2018c,eckenhoff2018closed} and keyframe-based odometry and SLAM systems~\cite{kerl2013dense,wang2017stereo} are also interesting future research directions. 

\section*{Acknowledgment}
This article solely reflects the opinions and conclusions of its authors and not TRI or any other Toyota entity. The authors would like to thank Andreas Girgensohn for the helpful discussion on choosing the HSV colormap.

\clearpage
\bibliographystyle{IEEEtran}
\IEEEtriggeratref{18}
\bibliography{strings-full,ieee-full,refs}

\end{document}